\def\year{2021}\relax
\documentclass[letterpaper]{article} % DO NOT CHANGE THIS
\usepackage{aaai21}  % DO NOT CHANGE THIS
\usepackage{times}  % DO NOT CHANGE THIS
\usepackage{helvet} % DO NOT CHANGE THIS
\usepackage{courier}  % DO NOT CHANGE THIS
\usepackage[hyphens]{url}  % DO NOT CHANGE THIS
\usepackage{graphicx} % DO NOT CHANGE THIS
\usepackage{graphicx}  % DO NOT CHANGE THIS
\usepackage{natbib}  % DO NOT CHANGE THIS AND DO NOT ADD ANY OPTIONS TO IT
\usepackage{caption} % DO NOT CHANGE THIS AND DO NOT ADD ANY OPTIONS TO IT
\usepackage{subfigure}
\usepackage{mathrsfs}
\usepackage{algorithm}
\usepackage{algorithmic}
\usepackage{booktabs}
\usepackage{multirow}

\title{AAG: Self-Supervised Representation Learning \\ 
	by Auxiliary Augmentation with GNT-Xent Loss}
\author{
	Yanlun Tu\textsuperscript{\rm 1}, Jianxing Feng\textsuperscript{\rm 2}, Yang Yang\textsuperscript{\rm 1*} \\
}
\affiliations{
 	\textsuperscript{\rm 1} Department of Computer Science and Engineering, \\
 	Shanghai Jiao Tong University, Shanghai, China. \\
 	\textsuperscript{\rm 2} Haohua Technology Co., Ltd, Shanghai, China. \\
 	tuyanlun@sjtu.edu.cn, fengjianxing@harmon.health, yangyang@cs.sjtu.edu.cn\\
}
\begin{document}
	
	\maketitle
	
	\begin{abstract}
		Self-supervised representation learning is an emerging research topic for its powerful capacity in learning with unlabeled data. As a mainstream self-supervised learning method, augmentation-based contrastive learning has achieved great success in various computer vision tasks that lack manual annotations. Despite current progress, the existing methods are often limited by extra cost on memory or storage, and their performance still has large room for improvement. Here we present a self-supervised representation learning method, namely AAG, which is featured by an auxiliary augmentation strategy and GNT-Xent loss. The auxiliary augmentation is able to promote the performance of contrastive learning by increasing the diversity of images. The proposed GNT-Xent loss enables a steady and fast training process and yields competitive accuracy. Experiment results demonstrate the superiority of AAG to previous state-of-the-art methods on CIFAR10, CIFAR100, and SVHN. Especially, AAG achieves 94.5$\%$ top-1 accuracy on CIFAR10 with batch size 64, which is 0.5\% higer than the best result of SimCLR with batch size 1024.
	\end{abstract}

	\section{Introduction}
	\noindent Image representation has always been the focus of computer vision studies. At the early developmental stage of image processing, images are represented by manually extracted global features \cite{jain1996image,manjunath1996texture}, including color, texture, and edge information. As these features are sensitive to variable lighting conditions and occlusion, they have been largely replaced by local feature extractors, like BoW \cite{sivic2003video} and SIFT \cite{lowe2004distinctive}, while the representation ability has still been limited by the hand-crafted extracting rules. 
	
	For the past decade, automatic feature learning by deep neural networks (DNNs) has dominated the field of image representation. DNNs learn feature embeddings through multiple layers of non-linear transformations. The output high-level abstract features generalize well in various downstream tasks. Especially, convolutional neural networks trained on images with annotated class labels can capture visual similarity among different categories to successfully make a semantic classification. However, due to the high cost or expertise required for manual annotation, a substantial ratio of image data remains unlabeled, which has not been effectively exploited. 
	
	In recent years, self-supervised learning methods \cite{jing2019self-supervised} have achieved breakthrough performance in the field of computer vision. The key to the success of these methods lies in the design of pretext tasks.  As a special form of unsupervised learning, the ``label'' in self-supervised learning is derived from the data itself. Proper pretext tasks can make full use of the inherent properties of data to improve the quality of learned representation and enhance the performance of downstream tasks. Typical pretext tasks include image generation \cite{goodfellow2014generative}, colorization \cite{zhang2016colorful}, inpainting \cite{pathak2016context}, and super resolution \cite{ledig2016photo-realistic}, which focus on specific operation but can also have good generalization ability on downstream tasks. Another kind of pretext tasks is context-based methods \cite{doersch2015unsupervised,noroozi2016unsupervised,gidaris2018unsupervised}, which utilizes spatial relations among different image patches or intrinsic attributes of images. 
	
	In this study, we focus on the task of instance-wise discrimination based on data augmentation, which is commonly-used in self-supervised learning. To create positive and negative labels, data augmentation techniques are utilized to generate views of images. The pairs of views originated from the same image form positive samples, and those from different images form negative samples. This kind of representative methods mainly includes batch-based and memory bank-based methods. A major limitation of these methods is the extra cost on storing the features of samples. For instance, the memory bank-based methods \cite{wu2018unsupervised,huang2019unsupervised,han2020a} allocate large memory space to maintain the features of generated views for all samples. MoCo \cite{he2019momentum} also requires to build a queue for feature storage, while the differences lie on the encoder for generating features of keys and the variable length of the queue. Instead, SimCLR \cite{chen2020a} directly fetches samples from batches. To ensure good performance, SimCLR has a much larger batch size, which restricts its applications in the labs of limited computational resources.

	Therefore, how to get a comparable or even better result with less resource consumption has become an attractive topic in the self-supervised learning field. Considering that the end-to-end methods are more straightforward and flexible compared to memory bank and momentum-based methods, we mainly focus on the batch-based methods. To address the above challenge, we propose AAG, i.e. self-supervised representation learning by \underline{A}uxiliary \underline{A}ugmentation with \underline{G}NT-Xent Loss. Specifically, to obtain recognizable features, a self-supervised pretext task based on instance discrimination is adopted. The main idea is to treat each image instance as a single ``class’’ so that the model has a direction for training. 
	
	The AAG model is driven by a hybrid data augmentation scheme using both basic and auxiliary data augmentation strategies, which generate three enhanced views for each image. Combinations of these views compose positive and negative samples that are fed into a siamese neural network to encode image features. Besides, to speed up the model training process while maintaining stability, we propose a simple but efficient contrastive loss function named GNT-Xent Loss. 
	Optimizing this modified contrastive loss enforces the positive pairs more closer while negative pairs more separated, thus yielding more discriminative representations of images. This method not only gets rid of memory bank but also uses a much smaller batch size compared to previous batch-based methods. 
	
	To assess the discriminant ability of the learned representation, we use both weighted $k$NN algorithm and linear evaluation to evaluate the model performance. 
	The contributions of our work are summarized as follows.
	\begin{itemize}
		\item We design a new scheme of contrastive learning with basic and auxiliary data augmentation. The hybrid data augmentation strategy greatly alleviates the dependency on large batch size.
		\item We propose a novel contrastive loss function which can not only maintain the stability of the training process but also improve the accuracy under both $k$NN and linear evaluation. %And we demonstrate why this loss function performs well on the benchmark with theoretical analysis and experiments.
		\item The AAG method achieves SOTA accuracies on multiple benchmark datasets with low computation cost.
	\end{itemize}
	
	\section{Related Work}
	\noindent In this section, we provide a brief overview of research progress on self-supervised representation learning in recent years, mainly involving contrastive learning.
	\subsubsection{Contrastive learning} uses contrastive loss function to measure the distance between samples \cite{hadsell2006dimensionality}. The main idea is to first learn the low-dimensional mapping of raw data, then decrease the Euclidean distance of similar pairs and increase that of dissimilar pairs. It allows us to retain the original semantics of samples even when the dimensions are greatly reduced. Contrastive learning is an effective method to learn the common representation of samples in different categories. 
	\subsubsection{Memory bank-based methods} compute the contrastive loss by using the image representation stored in the memory bank with features of current minibatch \cite{wu2018unsupervised,wu2018improving,he2019momentum,huang2019unsupervised,han2020a}. The memory bank itself will be updated iteratively during the training process. Wu et al. (2018) proposed a memory bank with appropriate parameter settings for the first time in the field of unsupervised learning. Then MoCo \cite{he2019momentum} replaces the memory bank with a variable length queue and achieves breakthrough results on a large-scale dataset. 
	\subsubsection{Batch-based methods} compute the contrastive loss in the current batch during the training process \cite{ye2019unsupervised,chen2020a}. The advantage of these methods is that the features for comparison are up-to-date at every time step. ISIF \cite{ye2019unsupervised} shows the superiority of batch-wise contrastive learning. SimCLR \cite{chen2020a} uses a quite large batch size to train the network with diversiform data augmentation approaches and achieves impressive results under the protocol of linear evaluation. 
	\begin{figure*}
		\centering
		\includegraphics[width=1\textwidth]{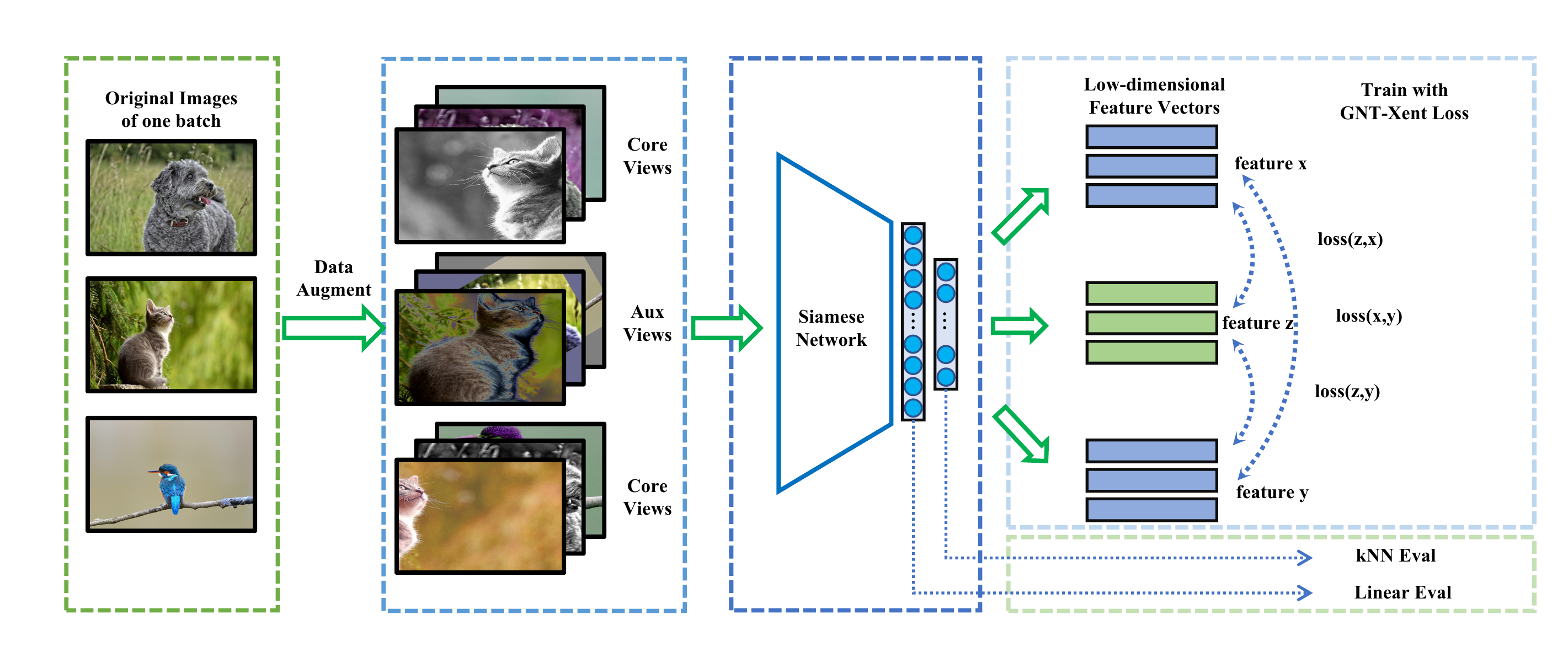}
		\caption{The framework of AAG. For one minibatch of each iteration in the training process, each image is enhanced with different data augmentation policies for three times, resulting into a triplet of views. These views are fed into a siamese network to extract low-dimensional feature vectors afterward. In the end, we calculate the GNT-Xent loss based on these features for optimization. We use both $k$NN and linear evaluation protocols to evaluate the learned representations.}
		\label{fig_overview}
	\end{figure*}

	\section{Methodology}
	\subsection{Model Overview}
	To fully exploit the potential of self-supervised learning in extracting features from images, we propose \textbf{A}uxiliary \textbf{A}ugment with \textbf{G}NT-Xent loss (\textbf{AAG}) method. %We follow the pattern of contrastive learning, but add an auxiliary view and modify the loss function on this basis. 
	The overview of our method is illustrated in Figure \ref{fig_overview}. Given a dataset of $N$ images $\mathcal{D}=\left\{ d_1,d_2,...,d_N \right\}$, our goal is to learn a function $v=f_\theta(d)$ without supervision, where $f_\theta(\cdot)$ is a deep neural network which maps image $d$ to feature $v$. During the training process, we randomly sample a minibatch of $n$ images at each iteration. For every image $d_i, i \in \left\{1,..,n\right\}$, we perform data augmentation for three times with basic augmentation and auxiliary augmentation. Three views are obtained by $core_{i,1}=g_{basic_1}(d_i)$, $core_{i,2}=g_{basic_2}(d_i)$ and $aux_i=g_{aux}(d_i)$, respectively, where $g_{basic_1}$ and $g_{basic_2}$ are sampled from the same family of basic augmentation while $g_{aux}$ is sampled from auxiliary augmentation. Then we feed these three views into the backbone network and get three feature embeddings, $x_i=f_\theta(core_{i,1}), y_i=f_\theta(core_{i,2}), z_i=f_\theta(aux_i)$. Finally, we calculate GNT-Xent Loss as defined in Eq. (\ref{eq:loss}). The loss is a sum of three components, each of which is the loss arising from differentiating a positive pair of views against the corresponding negative ones. The formal definitions of these three components are shown in Algorithm \ref{algorithm:overview}.
	\begin{equation}\label{eq:loss}
	\mathcal{L}_i=\mathcal{L}_{x_i y_i} + \mathcal{L}_{z_i x_i} + \mathcal{L}_{z_i y_i}
	\end{equation} 
	
	For a comprehensive assessment on the quality of learned feature embeddings, we use two common evaluation methods, i.e. weighted $k$NN \cite{wu2018unsupervised} and linear evaluation protocol \cite{bachman2019learning,chen2020a}.
	
	\subsection{Hybrid Data Augmentation Scheme}
	Data augmentation makes instance-wise discrimination a practical pretext task for self-supervised learning (SSL). As most of the downstream tasks depend on the discovery of high-level semantic meaning in images, a basic assumption of data augmentation-based SSL is that the semantic content of images is invariant to the augmentation operations. Thus, the basic goal of data augmentation is to bring diversification to image features and help the model distill the semantic information.

	\subsubsection{Basic data augmentation.}\label{sec:basicaug} Basic data augmentation is defined as the superposition of a series of random data augmentation approaches. Previous studies \cite{wu2018unsupervised,chen2020a} have shown that data augmentation is crucial for image preprocessing. Whether data augmentation is applied properly determines the final result of the trained model. Following the practice of previous batch-based methods \cite{chen2020a}, our basic data augmentation consists of the following 5 operations that are performed sequentially with random magnitudes.
	
	i) Random resize and crop;
	
	ii) Color jitter;
	
	iii) Random grayscale;
	
	iv) Random horizontal flip;
	
	v) Gaussian blur.

	%1) random resize and crop, 2) color jitter, 3) random grayscale, 4) random horizontal flip, and 5) gaussian blur. They are performed sequentially in present order with different magnitudes. The parameter configuration is referred to SimCLR. After an image is augmented twice with basic data augmentation, two enhanced views with different image attributes will be obtained, which are called \textit{core views}. 
	
	By calling the basic data augmentation on an image twice, we obtain two views with different image attributes, which are called \textit{core views}.
	
	\subsubsection{Auxiliary data augmentation.} The basic data augmentation adopts a conservative scheme covering only five basic operations, which may be unable to generate sufficiently diversified image features. Some previous methods address this issue by increasing the number of samples, thus resulting into a large batch size. To reduce the dependency on batch size, we propose the auxiliary data augmentation strategy to generate additional views. The purpose of using auxiliary data augmentation is to bring in more data augmentation operations with more randomness so that the trained model can learn the semantic information from images with various attributes. %Based on the additional augmentation operations, we generate a third view for each image, which is called
	
	Ideally, the more semantic-invariant transformations are included, the more easily models generalize well. %CMC \cite{tian2019contrastive} shows that more views lead to better representations that capture underlying scene semantics. 
	However, the model performance does not necessarily benefit from adding augmentation operations, as the augmentation operations are not guaranteed to be semantic-invariant. The larger the operation pool, the higher risk may be introduced and further result into noisy samples in the dataset. 

	Therefore, to obtain a relatively reliable operation set, we choose good policies demonstrated by previous data augmentation studies, mainly from AutoAugment \cite{cubuk2019autoaugment:} and RandAugment \cite{cubuk2019randaugment:}. AutoAugment \cite{cubuk2019autoaugment:} uses an adaptive search algorithm to find the best data augmentation policies for different datasets. The policy has several sub-policies. Each sub-policy consists of 2 operations which are associated with two values, i.e. the probability to use the operation and the magnitude of the operation. AutoAugment provides 25 best sub-policies for CIFAR10, SVHN, and ImageNet. The searched optimal policies can be used directly for the above datasets, while searching policies for a new dataset may be costly. RandAugment \cite{cubuk2019randaugment:} does not require a search process and only needs to define two parameters, namely the number of operations in a sub-policy and the magnitude of operations. It greatly reduces the time complexity of searching and achieves a close performance compared to AutoAugment. 
	
	Obviously, the operation pool for generating auxiliary views is much larger than that of basic augmentation. Therefore, to control the risk of introducing too much noise, we generate only one view for each image, which is called \textit{auxiliary view}. In this way, there is no positive pair consisting of two auxiliary views.% in case that two auxiliary views differ too much.
	
	%To summarize, we adopt a hybrid data augmentation scheme, including both basic augmentation and auxiliary augmentation, which generate two basic views and one auxiliary view, respectively, aiming to achieve a tradeoff between multiplicity of views and controlled random noise.
	
	\subsection{Pretext Task}
	\subsubsection{Instance discrimination with three views.} Instance discrimination \cite{wu2018unsupervised} forces the model to learn to recognize different instances rather than different classes in the absence of any semantic labels. As a consequence, representations that capture the similarity and differences between instances can be learned. Although it is a task for discriminating instances, semantic insights can also be learned. %This is the main idea of instance discrimination.
	%The contrastive learning method can naturally solve the problem of instance discrimination. When it is applied with three views, the calculating steps are as follows.
	Given three views for each image, the process of contrastive learning is as follows. 
	
	Suppose we have a minibatch of $n$ images at some time step. The L2-normalized feature embeddings of the $i$th image are $x_i$, $y_i$, and $z_i$ ($i \in \left\{1,..,n\right\}$), respectively, where $x_i$ and $y_i$ are derived from basic views, and $z_i$ is derived from the auxiliary view. As we use cosine similarity to measure the distance between feature vectors, it is essential to apply L2norm to scale the length of vector to 1. Let $s(v_1,v_2)=v_1^{\top}v_2/\left\|v_1\right\|\left\|v_2\right\|$ denote cosine similarity between two feature vectors $v_1$ and $v_2$, then the value of $s(v_1,v_2)$ ranges from -1 to 1. In order to scale up the range of similarity, we follow \cite{wu2018unsupervised} to apply a temperature parameter $\tau$. With the above parameters, we propose a new loss function termed as GNT-Xent (the gradient-stabilized and normalized temperature-scaled cross-entropy loss), which is inspired by the name of NT-Xent \cite{chen2020a}. The GNT-Xent loss for a positive pair $(x_i,y_i)$ is formulated as,
	\begin{equation}
	\mathcal{L}_{x_i y_i}=-\log \frac{\exp (s^+_i/\tau)} { \sum_{k=1}^N{_{k \neq i}} \exp (s^-_k/\tau)},
	\label{eq:gnt-xnet}
	\end{equation}
	where $s^+_i=s(x_i, y_i)$ and $s^-_k=s(x_k,y_i)$ denote the similarity of a positive pair and negative pair, respectively (formal definitions are in Algorithm \ref{algorithm:overview}). The main difference from NT-Xent is that $s^+$ is removed from the denominator. Detailed explanations are given in the next section. 
	
	%After the three losses is calculated, we use the sum of them to optimize the siamese neural network. Algorithm \ref{algorithm:overview} summarizes our method in detail. 
	
	\begin{algorithm}[t]
		\caption{the overall learning algorithm of AAG.}
		\hspace*{0.1in} {\bf Input:} %算法的输入， \hspace*{0.02in}用来控制位置，同时利用 \\ 进行换行
		batch size $n$, dataset $\mathcal{D}$, temperature $\tau$.\\
		\hspace*{0.1in} {\bf Output:} %算法的结果输出
		network $f_\theta$.
		\begin{algorithmic}
			\FOR{each minibatch $\left\{ d_i \right\}^n_{i=1} \in \mathcal{D}$}
			\FORALL {$i \in \left\{ 1,...,n \right\}$}
			\STATE Draw augmentataion functions 
			\STATE $g_{basic_1},g_{basic_2} \sim \mathcal{G}_{basic}$ and $g_{aux} \sim \mathcal{G}_{aux}$
			\STATE $ x_i=f_\theta(g_{basic_1}(d_i)) $
			\STATE $ y_i=f_\theta(g_{basic_2}(d_i)) $
			\STATE $ z_i=f_\theta(g_{aux}(d_i)) $
			\STATE Apply L2 norm to $x_i,y_i,z_i$
			\ENDFOR
			\FORALL {$i \in \left\{ 1,...,n \right\}$, $j \in \left\{ 1,...,n \right\}$, $j \neq i$}
			\STATE $s^+_{x_iy_i}=s(x_i,y_i)/\tau$
			\STATE $s^+_{z_ix_i}=s(z_i,x_i)/\tau$
			\STATE $s^+_{z_iy_i}=s(z_i,y_i)/\tau$
			\STATE $s^-_{x_ix_j}=s(x_i,x_j)/\tau, s^-_{y_iy_j}=s(y_i,y_j)/\tau$
			\STATE $s^-_{x_iy_j}=s(x_i,y_j)/\tau, s^-_{x_jy_i} =s(x_j,y_i)/\tau$
			\STATE $s^-_{z_ix_j}=s(z_i,x_j)/\tau, s^-_{z_jx_i} =s(z_j,x_i)/\tau$
			\STATE $s^-_{z_iy_j}=s(z_i,y_j)/\tau, s^-_{z_jy_i} =s(z_j,y_i)/\tau$
			\STATE $\mathcal{L}_{x_i y_i}=-\log \frac{e^{s^+_{x_iy_i}}} { \sum_{j}e^{ s^-_{x_iy_j}}+e^{s^-_{x_jy_i}}+e^{s^-_{x_ix_j}}+e^{s^-_{y_iy_j}}}$
			\STATE $\mathcal{L}_{z_i x_i}=-\log \frac{e^{ s^+_{z_ix_i}}} { \sum_{j}e^{ s^-_{z_ix_j}}}\cdot \frac{e^ {s^+_{z_ix_i}}} { \sum_{j} e^{s^-_{z_jx_i}}}$
			\STATE $\mathcal{L}_{z_i y_i}=-\log \frac{e^{ s^+_{z_iy_i}}} { \sum_{j}e^{s^-_{z_iy_j}}}\cdot \frac{e^ {s^+_{z_iy_i}}} { \sum_{j} e^{s^-_{z_jy_i}}}$
			\ENDFOR
			%\FORALL {$i \in \left\{ 1,...,n \right\}$}
			\STATE $\mathcal{L}=\frac{1}{n}\sum_{i=1}^{n}(\mathcal{L}_{x_i y_i}+\mathcal{L}_{z_i x_i}+\mathcal{L}_{z_i y_i})$
			%\ENDFOR
			\STATE Update siamese network $f_\theta$ to minimize $\mathcal{L}$
			\ENDFOR
			\STATE Return the network $f_\theta$
		\end{algorithmic}
		\label{algorithm:overview}
	\end{algorithm}
	
	\subsubsection{Design considerations of loss function.}\label{sec:loss}
	%	\subsubsection{Modification of NT-Xent.} 
	In this section, we will give a formal explanation on the reason for modifying NT-Xent. The original NT-Xent loss can be formulated as,
	\begin{equation}
	\mathcal{L}_{NT-Xent}=-\log \frac{\exp (s^+_i)} {\exp(s^+_i) + \sum_{k=1}^N{_{k!=i}} \exp (s^-_k)},
	\end{equation}
	where $s^+_i$ and $s^-_k$ denote the similarity between a positive pair and a negative pair respectively. The loss above aims to minimize $s^-_k$ and maximize $s^+_i$. Thus the limit value of loss function approaches 0 on the right side. Actually, this limit is not required for contrastive learning, becuase contrastive learning only focuses on assimilating positive pairs while dissimilating negative pairs.
	
	Based on this consideration, we propose a modified contrastive loss, GNT-Xent, as formulated in Eq. (\ref{eq:gnt}),
	\begin{equation}\label{eq:gnt}
	\mathcal{L}_{GNT-Xent}=-\log \frac{\exp (s^+_i)} {\sum_{k=1}^N{_{k!=i}} \exp (s^-_k)},
	\end{equation}
	
	As can be seen, the difference is that we subtract the item of the positive pair from the denominator. Let's ignore the constant parameter $\tau$ and compute the gradients. 
	$\frac{\partial \mathcal{L}_{NT-Xent}}{\partial s_i^+} = - \frac{\sum_{k=1}^N{_{k\neq i}} \exp (s^-_k)}{\exp(s_i^+)+\sum_{k=1}^N{_{k\neq i}} \exp (s^-_k)}$ and $\frac{\partial \mathcal{L}_{NT-Xent}}{\partial s_k^-} = \frac{\exp(s_k^-)}{\exp(s_i^+)+\sum_{k=1}^N{_{k\neq i}} \exp (s^-_k)}$. During the training , the value of $s_i^+$ keeps increasing rapidly, so gradients of $\frac{\partial \mathcal{L}_{NT-Xent}}{\partial s_i^+}$, $\frac{\partial \mathcal{L}_{NT-Xent}}{\partial s_k^-}$ will be affected and reduced. As a result, the training process can be hampered in the later stages. By contrast, the gradients of GNT-Xent will not be affected by the values of $s_i^+$ or $s_i^-$, because they are constant. It is trivial to deduce that $\frac{\partial \mathcal{L}_{GNT-Xent}} {\partial s_i^+}=-1$ and 
	$\frac{\partial L_{GNT-Xent}}{\partial s_k^-} = \frac{exp(s_k^-)}{\sum_{k=1,k \neq i}^{N} exp(s_k^-)}$. The sum of $\frac{\partial L_{GNT-Xent}}{\partial s_k^-}$ is $1$. Then the training process can be steady and continuous. %And our experiments show that our modification is reliable and effective.
	
	%	\subsubsection{Design of total loss.} 
	The total loss of AAG consists of three components. As three positive pairs can be derived from three views for each image, each component aims to penalize for the prediction error of a positive pair. Here we have different treatments regarding the negative samples when calculating the losses. The major difference is that in the first component, we consider all the core-core pairs of views from different images; while in the second and third components, although the positive pairs contain an auxiliary view, we do not take auxiliary-auxiliary pairs of views from different images into consideration. The major reason is that the auxiliary augmentation has much randomness and has more chances to produce noisy samples compared to basic augmentation.%lt into two auxiliary views from two images are mostly very dissimilar, thus inclusion of such pairs has little aid for training the model to recognize semantic meaningful patterns.

	\section{Experiments}
	\subsubsection{Experimental settings.}
	To assess the performance of AAG working with different models, we experiment with three backbone networks, namely AlexNet \cite{krizhevsky2012imagenet}, ResNet18, and ResNet50 \cite{he2016deep}. For training the model, we set the batch size to 128 and the number of epochs to 200, and use SGD with momentum. The weight decay parameter is 5$\times 10^{-4}$ and momentum is set to 0.9. The embedding feature size of the last layer is 128. The initial learning rate is 0.03. We replace the StepLR schedule with a CosineLR schedule \cite{loshchilov2016sgdr} without restarts. In the GNT-Xent loss, we set the temperature parameter $\tau$ to 0.1 as suggested in \cite{wu2018unsupervised}. In the experiments except the ablation study, we adopt the policies provided by AutoAugment as the default auxiliary augmentation polices. Most of the experiments were implemented in pytorch running on GeForce RTX 2080 Ti. And the experiments with large batch size were performed in Ascend 910 processor on Huawei Cloud.

	\subsubsection{Evaluation protocols.}
	We adopt two common methods to evaluate the performance of self-supervised learning, namely $k$NN evaluation \cite{cover1967nearest,wu2018unsupervised} and linear evaluation \cite{zhang2016colorful,bachman2019learning,chen2020a}. The $k$NN evaluation computes low-dim features of the last layer and compares them against the ones of training images in the memory bank, using cosine similarity. The top $k$ nearest neighbors will be used to make the prediction. Linear evaluation fixes parameters of the trained network and utilizes the features before the last layer to retrain a one-layer network. The accuracy of the one-layer network is regarded as the result of linear evaluation. We use Adam optimizer with the initial learning rate of 0.01 to train the one-layer network for 50 epochs. The CosineLR schedule is also applied.
	\begin{table*}[!htb]
		\centering
		\caption{$k$NN evaluation on different methods and datasets. Results marked as * are borrowed from previous work \cite{ye2019unsupervised,huang2019unsupervised,han2020a}}
		\label{tab_$k$NN_200}
		\begin{tabular}{lccccccc}
			\toprule
			Dataset & \multicolumn{2}c{CIFAR10} & \multicolumn{2}c{CIFAR100} & \multicolumn{2}c{SVHN} & \multirow{2}{*}{Memory Bank}\\
			(Network) & ResNet18 & AlexNet & ResNet18 & AlexNet & ResNet18 & AlexNet & \\
			\midrule
			DeepCluster* & 67.6 & 62.3 & - & 22.7 & - & 84.9 & - \\
			Instance* & 80.8 & 60.3 & 50.7 & 32.7 & 93.6 & 79.8 & $\surd$\\
			ISIF* & 83.6 & 74.4 & 54.4 & \textbf{44.1} & 91.3 & 89.8 & $\times$\\
			SimCLR* & 82.3 & 73.0 & 55.8 & 43.2 & 90.8 & 88.6 & $\times$\\
			AND* & 86.3 & 74.8 & 57.2 & 41.5 & 94.4 & 90.9 & $\surd$\\
			AAG (Ours) & \textbf{88.3} & \textbf{74.9} & \textbf{60.6} & 39.7 & \textbf{95.3} & \textbf{92.6} & $\times$\\
			\bottomrule
		\end{tabular}
	\end{table*}
	
	\begin{table*}
		\centering
		\caption{The results of AAG using both $k$NN and linear evaluation and Super-AND using $k$NN evaluation with 1000 training epochs.}
		\label{tab_$k$NN_linear_1000}
		\begin{tabular}{lccccccc}
			\toprule
			Dataset & \multicolumn{2}c{CIFAR10} & \multicolumn{2}c{CIFAR100} & \multicolumn{2}c{SVHN} \\
			(Network) & ResNet18 & AlexNet & ResNet18 & AlexNet & ResNet18 & AlexNet & \\
			\midrule
			Super-AND ($k$NN Eval) & 89.2 & 75.6 & 61.5 & 42.7 & 94.9 & 91.9 \\
			AAG ($k$NN Eval) & \textbf{91.2} & \textbf{81.2} & \textbf{64.9} & \textbf{50.0} & \textbf{95.6} & \textbf{93.0} \\
			AAG (Linear Eval) & \textbf{91.6} & \textbf{81.2} & \textbf{66.4} & \textbf{51.2} & \textbf{96.3} & \textbf{94.3} \\
			\bottomrule
		\end{tabular}
	\end{table*}
	\begin{table}
		\centering
		\caption{Comparison between Super-AND and AAG on CIFAR10 in a single training process of 1000 epochs under $k$NN evaluation.}
		\label{tab_cmp_1000}
		\begin{tabular}{lccccc}
			\toprule
			
			Training epochs & 200 & 400 & 600 & 800 & 1000 \\
			\midrule
			Super-AND & 84.8 & \textbf{87.4} & 88.2 & 89.1& 89.2 \\
			AAG & 84.8 & 86.9 & \textbf{89.4} & \textbf{90.3} & \textbf{91.2} \\
			\bottomrule
		\end{tabular}
	\end{table}
	\begin{table}
		\centering
		\caption{Comparison between NT-Xent and GNT-Xent on CIFAR10 under $k$NN evaluation.}
		\label{tab_loss_cmp}
		\begin{tabular}{lcc}
			\toprule
			Loss & $lr=0.03$ & $lr=0.3$ \\
			\midrule
			NT-Xent & 86.8 & 84.2 \\
			GNT-Xent & \textbf{88.3} & \textbf{86.1} \\
			\bottomrule
		\end{tabular}
	\end{table}
	\begin{table}
		\centering
		\caption{Comparison between NT-Xent and GNT-Xent on CIFAR10 with ISIF and SimCLR respectively.}
		\label{tab_loss_cmp_isif_simclr}
		\begin{tabular}{lcccc}
			\toprule
			Method & Evaluation & NT-Xent &GNT-Xent \\
			\midrule
			ISIF & $k$NN & 83.3 & \textbf{85.5} \\
			SimCLR & Linear & 82.9 & \textbf{85.8} \\
			\bottomrule
		\end{tabular}
	\end{table}
	\begin{table}
		\centering
		\caption{Comparison with SimCLR and AMDIM on CIFAR10.}
		\label{tab_resnet50_cmp}
		\begin{tabular}{lcccc}
			\toprule
			Method & Network & Batch Size & Linear Eval \\
			\midrule
			AMDIM & ResNet50 (25$\times$) & - & 91.2 \\
			SimCLR & ResNet50 (MLP) & 1024 & 94.0 \\
			AAG & ResNet50 (MLP) & 64 & \textbf{94.5} \\
			\bottomrule
		\end{tabular}
	\end{table}
	\subsubsection{Datasets.}
	\textbf{CIFAR10} \cite{cifar} is a natural image dataset which contains 60000 color images of size $32\times 32$ and 10 classes, among which 50000 images are for training and 10000 for testing. The image size of \textbf{CIFAR100} \cite{cifar} is the same as in CIFAR10, while CIFAR100 has 100 classes and each class has 600 images. In the \textbf{SVHN} dataset \cite{svhn}, images were cropped patches of house numbers from the Google Street View images, which has 73257 images for training and 26032 for testing. The number of classes is 10 and the size of images is also $32\times 32$.
	
	\subsection{Results and Discussions}
	\subsubsection{Baselines.} We adopt six baseline models for comparison, including DeepCluster \cite{caron2018deep}, Instance \cite{wu2018unsupervised}, ISIF \cite{ye2019unsupervised}, AND \cite{huang2019unsupervised}, Super-AND \cite{han2020a}, and SimCLR \cite{chen2020a}.
	
	\subsubsection{Evaluation for short-term training.} Table~\ref{tab_$k$NN_200} describes the $k$NN evaluation on six models including the proposed AAG while excluding Super-AND, because Super-AND was trained for 5 rounds with 200 epochs per round (i.e. a total of 1000 epochs which is five times ours). For a fair comparison with Super-AND, we conduct another experiment shown in Table~\ref{tab_$k$NN_linear_1000}. From Table~\ref{tab_$k$NN_200}, we can see that without the memory bank, AAG still outperforms all baselines except for one case. And we find that by using a complex CNN, the performance of AAG can get more benefits. As a result, without using a memory bank, our method outperforms the state-of-the-art methods under the same condition with lower computation complexity. 
	\subsubsection{Evaluation for long-term training.} Experimental results show that AAG is far from convergence under the condition of 200 epochs. Table~\ref{tab_cmp_1000} shows that Super-AND performs a little bit better in the early stage but AAG achieves higher accuracy by a large margin in the later stage. It suggests that the accuracy of AAG can be further improved by more training epochs. Thus, we retrain our AAG for 1000 epochs on the above datasets. The results are shown in Table~\ref{tab_$k$NN_linear_1000}. AAG performs well with a longer training time under both $k$NN and linear evaluation. 
	
	Moreover, we experiment with different numbers of training epochs to investigate the impact of epoch number on the accuracy of $k$NN and linear evaluation on CIFAR10. As Figure~\ref{fig_epoch} shows, the accuracy is improved as the number of epochs increases, and the accuracy of linear evaluation is always higher than that of $k$NN evaluation. 
	
	\subsubsection{Investigation on batch size.} 
\begin{figure}[t]
		\centering
		\begin{minipage}[t]{0.22\textwidth}
			\centering
			\includegraphics[width=1\textwidth]{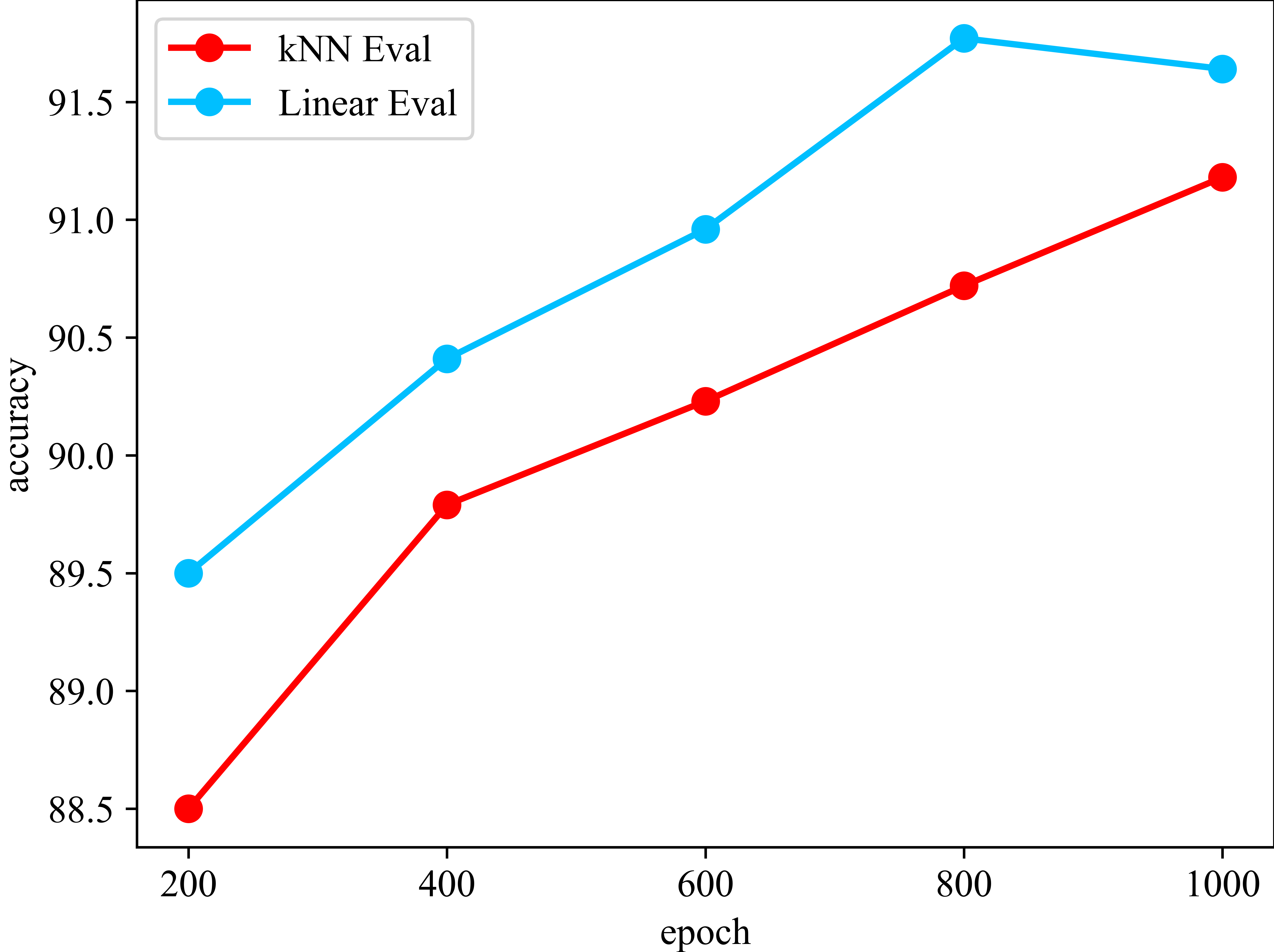}
			\caption{Impact of training epoch on CIFAR10.}
			\label{fig_epoch}
		\end{minipage}
		\hfill
		\begin{minipage}[t]{0.22\textwidth}
			\centering
			\includegraphics[width=1\textwidth]{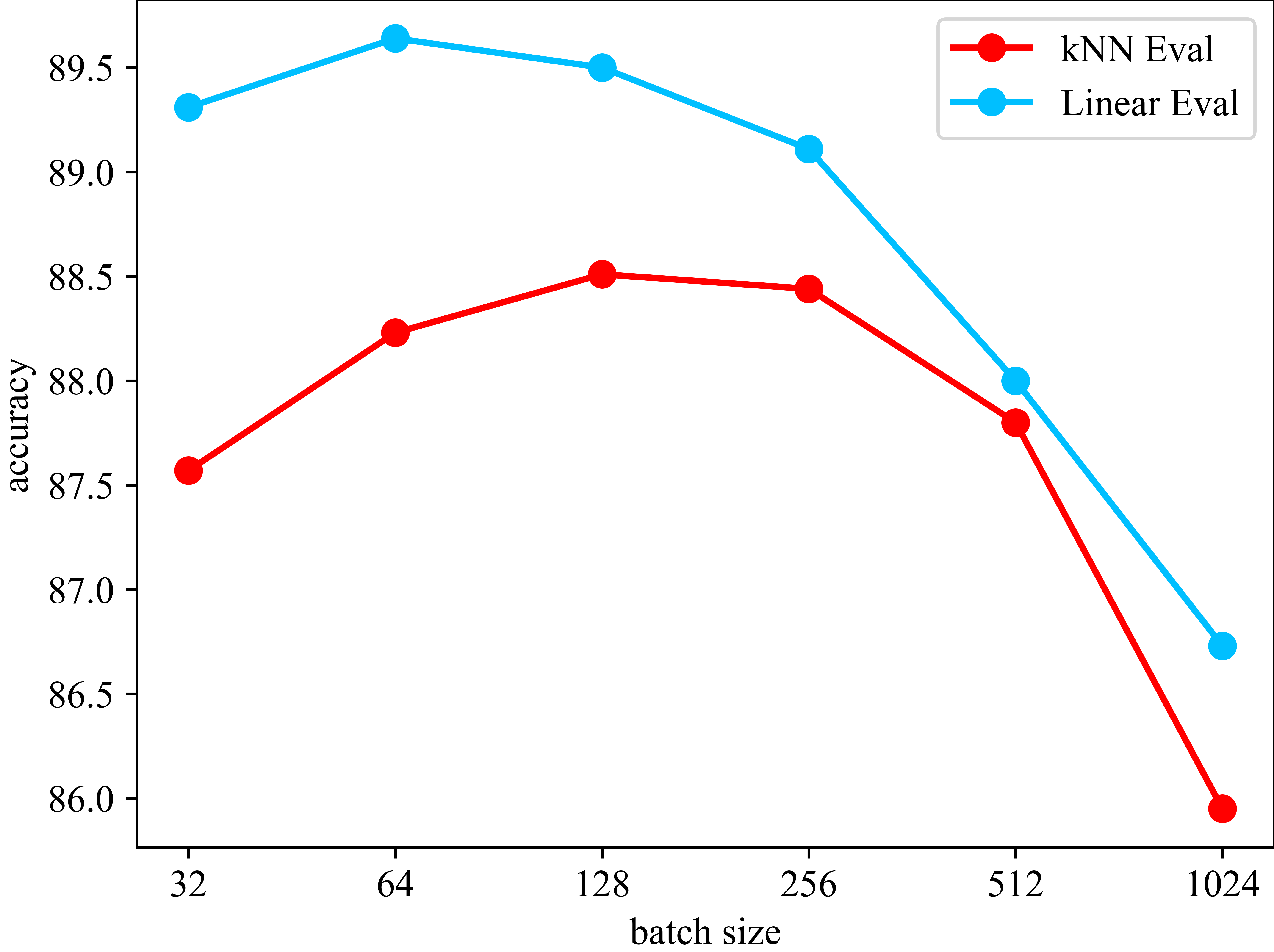}
			\caption{Impact of batch size on CIFAR10.}
			\label{fig_batchsize}
		\end{minipage}
	\end{figure}

To examine the effect of batch size in AAG, we conduct an experiment on CIFAR10 with varying batch size (Figure~\ref{fig_batchsize}). We use $Batch Size/128\times 0.03$ as the initial learning rate for the training of different batch sizes \cite{goyal2017accurate}. The training epoch is set to 200. For batch sizes more than 256, we warm up the learning rate for 10 epochs. The curve shows that the optimal batch size is not the largest one. The possible reason is that when the batch size gets larger, the positive samples being misclassified as negative pairs in a batch increases. Consequently, the overall performance declines.
	%\subsubsection{Feature Visualization.} To test the quality of representation learning, we visualize the 128D
	\subsubsection{Training process.} 
\begin{figure}[t]
		\centering
		\subfigure[Impact of gradients on $s^+$]{
			\begin{minipage}{0.22\textwidth}
				\includegraphics[width=1\textwidth]{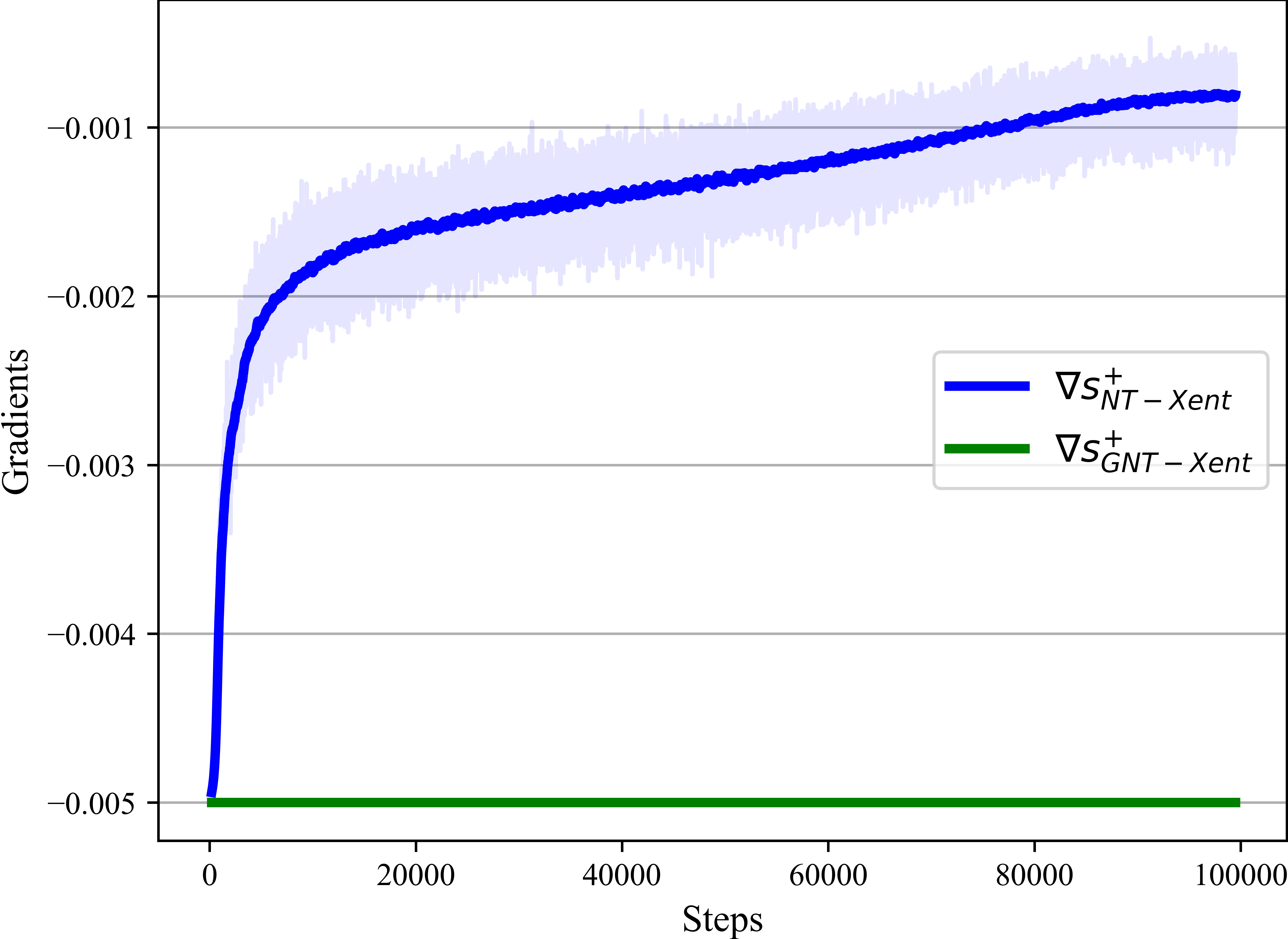}
			\end{minipage}
		}
		\subfigure[Impact of gradients on $s^-$]{
			\begin{minipage}{0.22\textwidth}
				\includegraphics[width=1\textwidth]{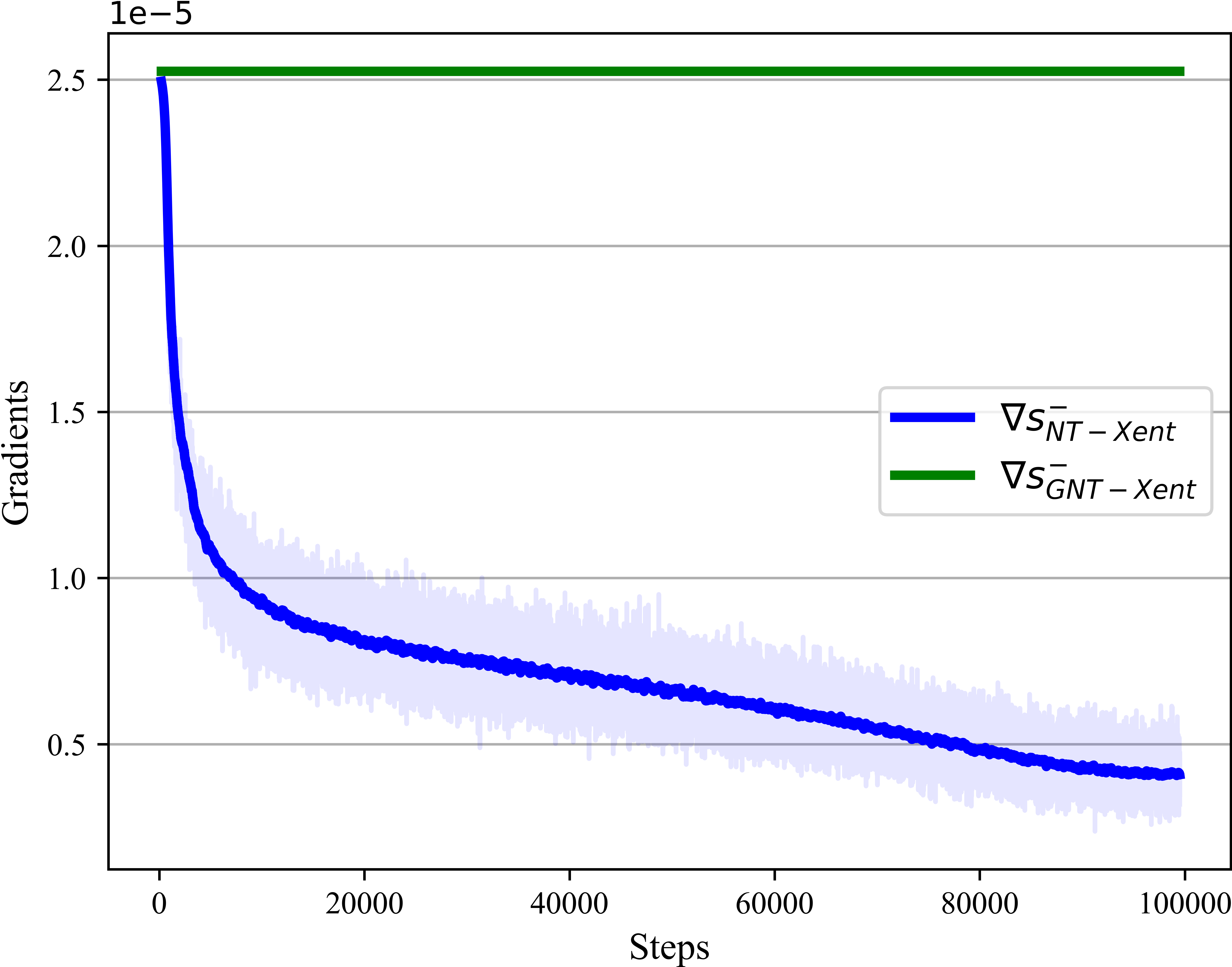}
			\end{minipage}
		}
		\subfigure[Impact of similarity on $s^+$]{
			\begin{minipage}{0.22\textwidth}
				\includegraphics[width=1\textwidth]{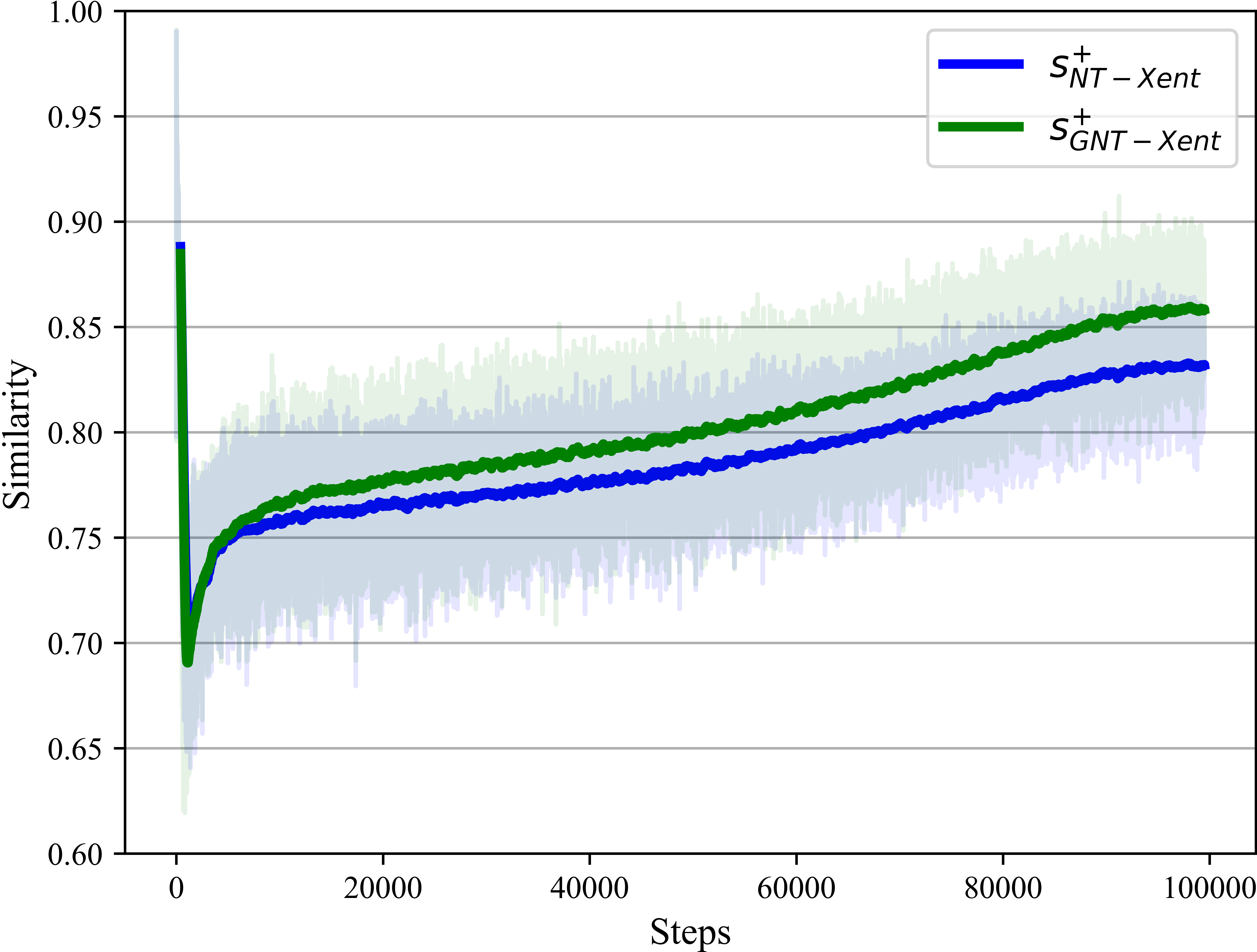}
			\end{minipage}
		}
		\subfigure[Impact of similarity on $s^-$]{
			\begin{minipage}{0.22\textwidth}
				\includegraphics[width=1\textwidth]{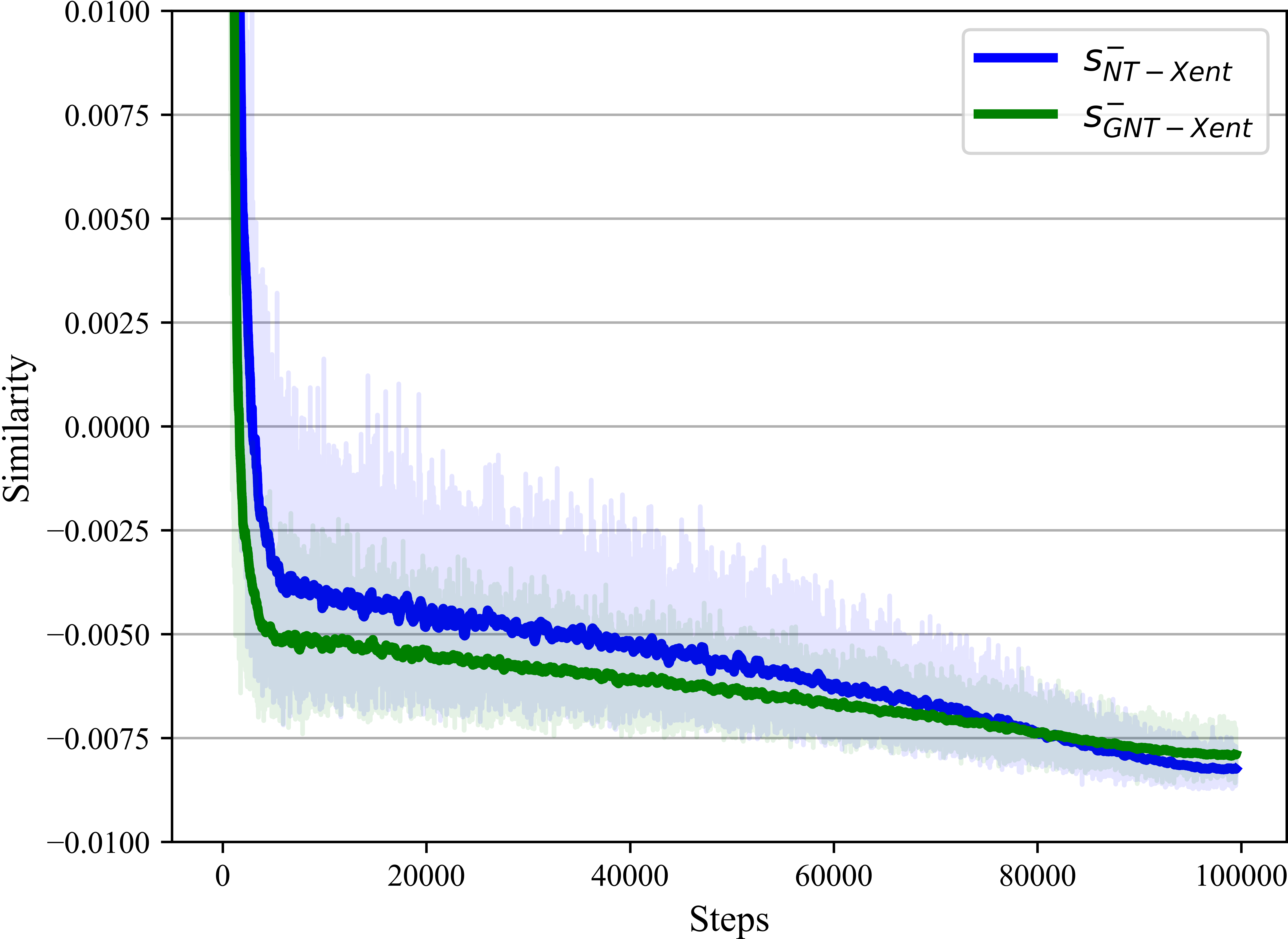}
			\end{minipage}
		}
		\caption{Comparison between NT-Xent and GNT-Xent on the values of $\nabla s^+$, $\nabla s^-$, $s^+$, and $s^-$ at each iteration (Curve smoothing is applied).}
		\label{fig_loss_cmp}
	\end{figure}

To verify our theoretical analysis on GNT-Xent, we conduct a comparison experiment between GNT-Xent and NT-Xent (results shown in Figure~\ref{fig_loss_cmp}). We use the same settings for both loss functions. As the values of gradients are influenced by the batch size, what we care about is the trend of changes rather than the specific value.
	
	As can be seen, $\nabla s^+$ and $\nabla s^-$ of GNT-Xent are constants during the training, whereas the values of NT-Xent decline sharply at the beginning and then going down steadily. As a result, the cosine similarity $s^+$ of GNT-Xent is higher than that of NT-Xent and the gap always exists. Furthermore, GNT-Xent achieves higher accuracy compared to NT-Xent. The results are reported in Table~\ref{tab_loss_cmp}. Besides, to prove that GNT-Xent does not simply speed up the learning process, we use a larger learning rate of 0.3 to retrain the model. Even though the performance of both loss functions gets worse, GNT-Xent still performs better than NT-Xent. 
	
	To further examine the efficacy of the proposed loss function, we conduct another comparison experiment with two existing methods \cite{ye2019unsupervised,chen2020a} which use NT-Xent as their loss function. As shown in Table~\ref{tab_loss_cmp_isif_simclr}, we replace their loss functions by GNT-Xent. The modified versions outperform original ones by a large margin. 
	
\begin{figure}[t]
		\centering
		\subfigure[accuracy]{
			\begin{minipage}{0.2\textwidth}
				\includegraphics[width=1\textwidth]{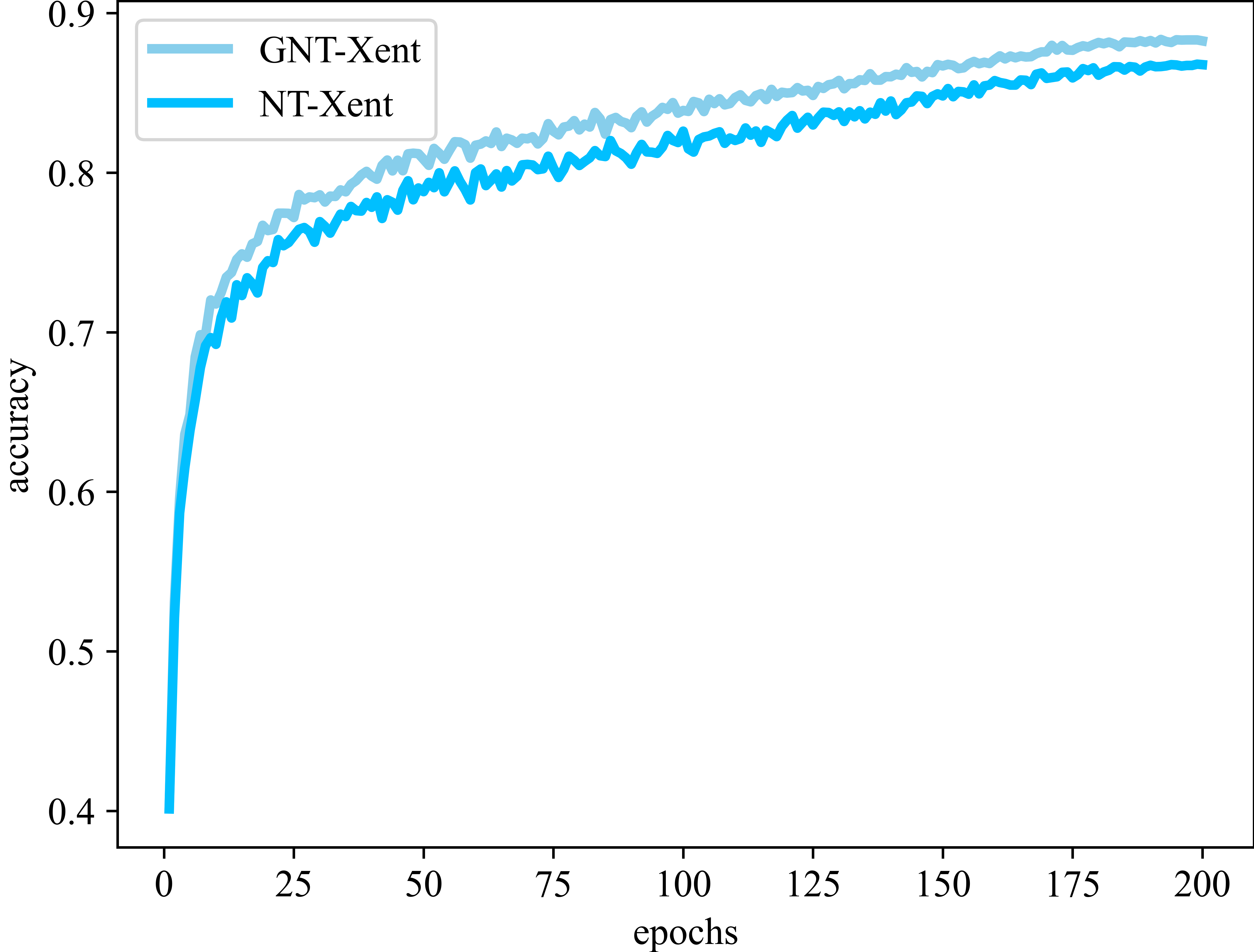}
			\end{minipage}
		}
		\subfigure[loss]{
			\begin{minipage}{0.22\textwidth}
				\includegraphics[width=1\textwidth]{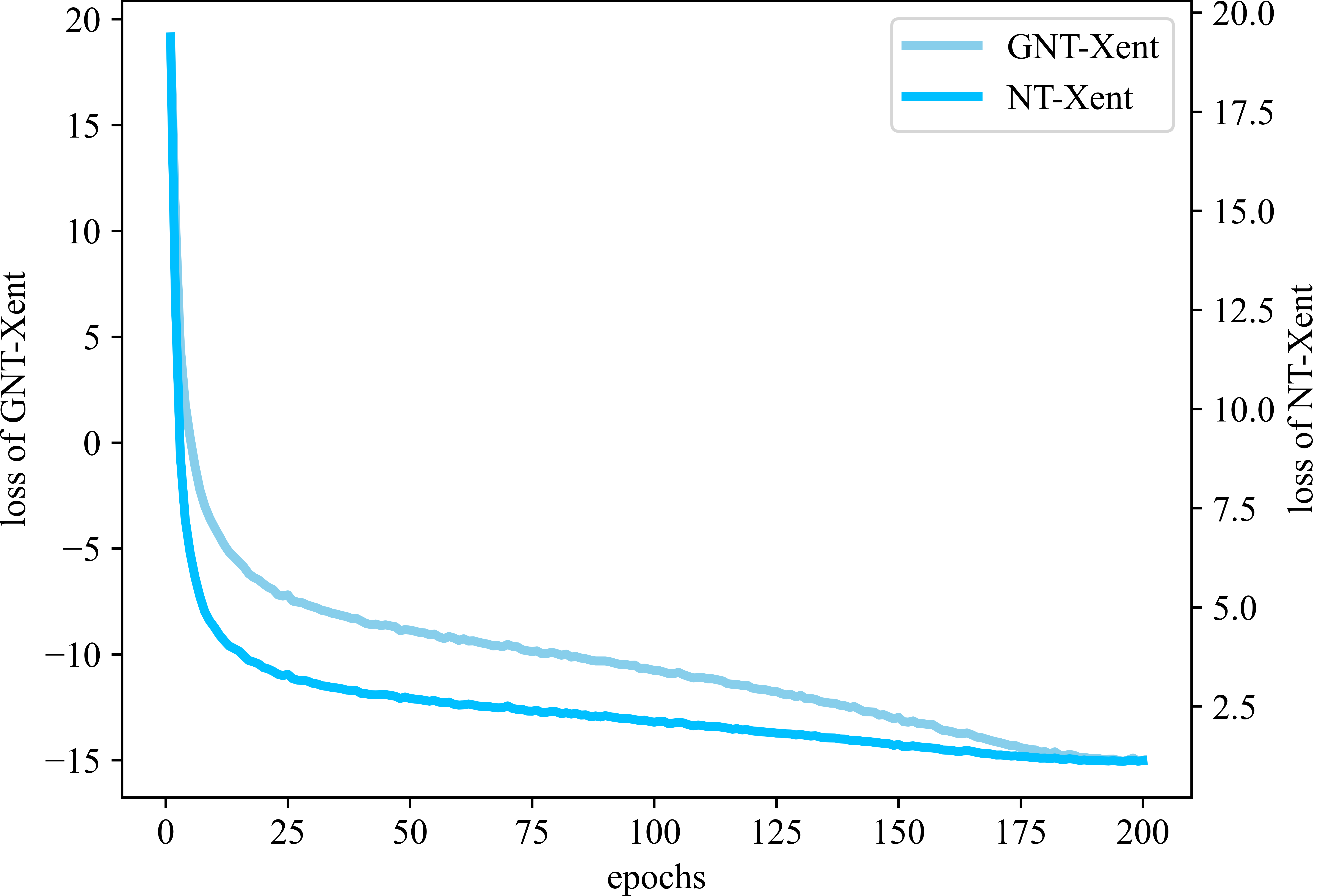}
			\end{minipage}
		}
		\caption{Curves of $k$NN accuracy and loss on CIFAR10 with different loss functions.}
		\label{fig_process_cmp}
	\end{figure}

\begin{figure}
		\centering
		\subfigure[NT-Xent] {
			\begin{minipage}{0.2\textwidth}
				\includegraphics[width=1\textwidth]{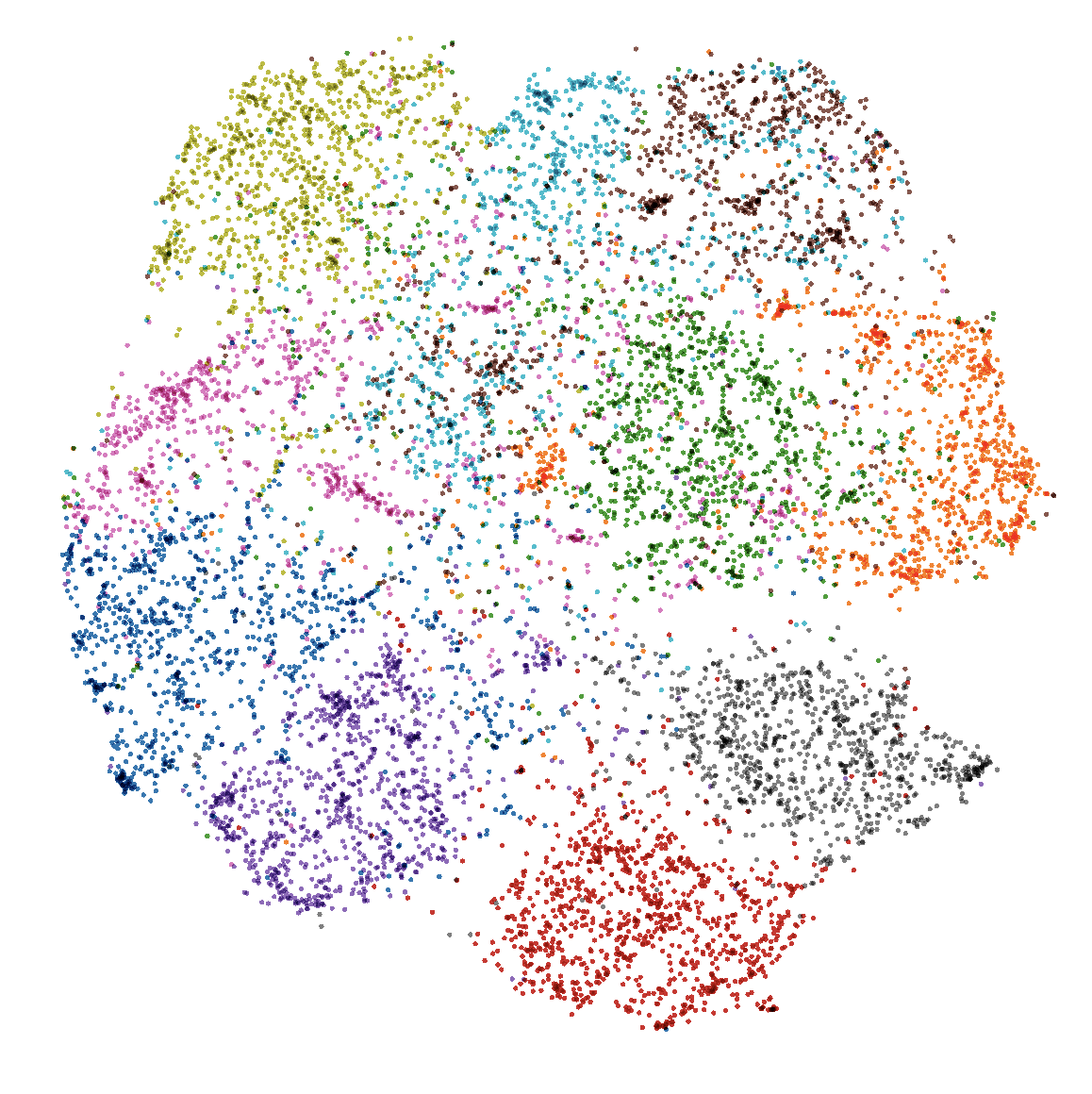}
			\end{minipage}
		}
		\subfigure[GNT-Xent] {
			\begin{minipage}{0.2\textwidth}
				\includegraphics[width=1\textwidth]{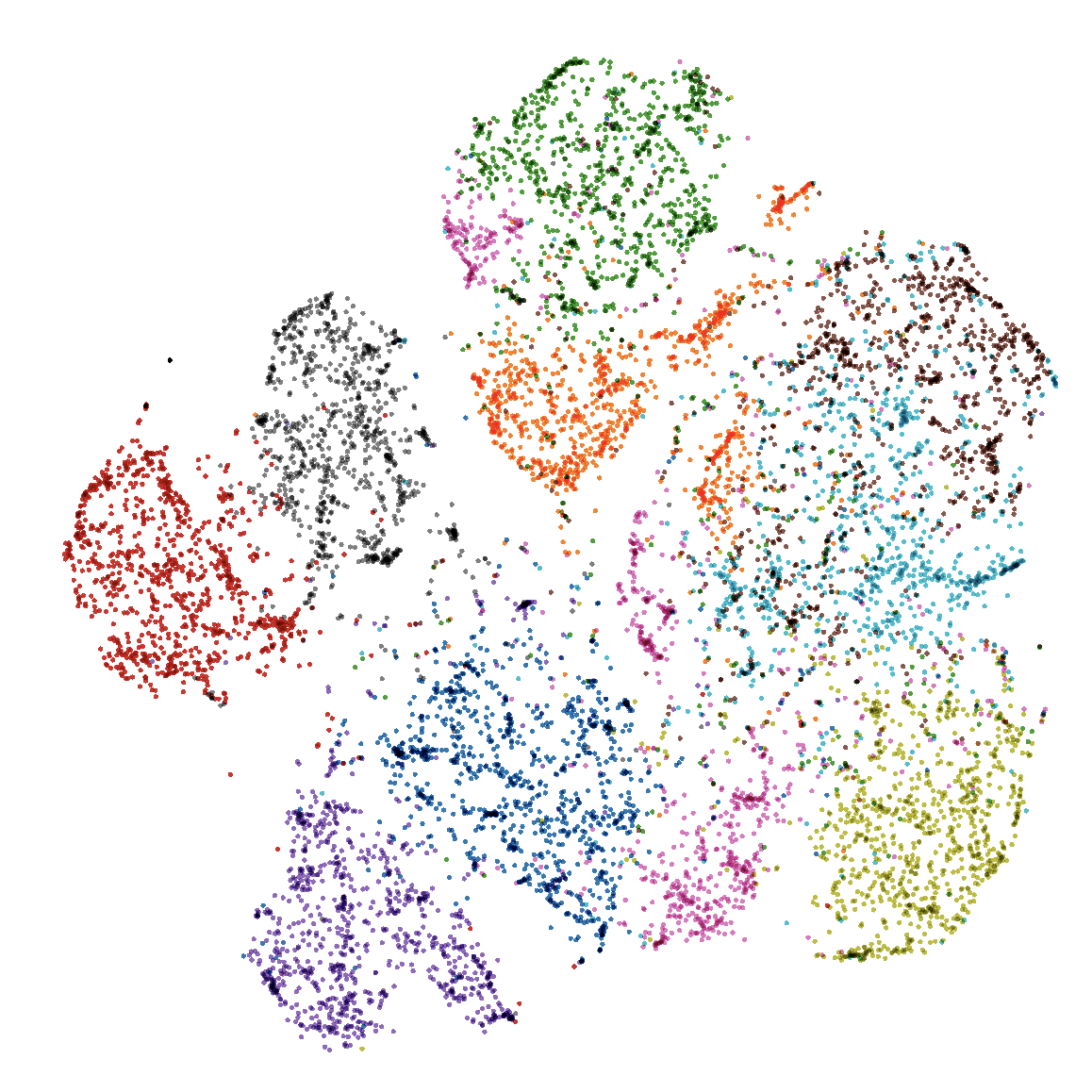}
			\end{minipage}
		}
		\caption{Visualization of features obtained by NT-Xent and GNT-Xent on CIFAR10.}
		\label{fig_tSNE}
	\end{figure}

	In addition, to visualize the discriminant capacity of learned features, we project the 128D features onto 2D space via the t-SNE algorithm \cite{maaten2008visualizing}. As Figure~\ref{fig_tSNE} shows, the features of GNT-Xent has a more separated distribution than that of NT-Xent. The results shown in Figure~\ref{fig_process_cmp}, i.e. the curves of $k$NN accuracy and loss during the training, again demonstrate the advantages of GNT-Xent over NT-Xent.

	\subsubsection{Train with a bigger network.} The SOTA method AMDIM \cite{bachman2019learning} achieves an accuracy of 91.2$\%$ on CIFAR10 with ResNet50 (25$\times$), while SimCLR outperforms AMDIM by using a larger batch size and longer training steps with a standard ResNet50. To compare with SimCLR, we use the same network with a 2-layer MLP as SimCLR does and train for 1000 epochs. Note that different from SimCLR, we use a batch size of 64 which allows training on a single GPU like RTX 2080Ti. Table ~\ref{tab_resnet50_cmp} shows that our method outperforms SimCLR's best result with batch size 1024. This result suggests that the batch size can be effectively reduced by using our method.
	\subsubsection{Ablation study.} To verify the validity of our method and investigate the impact of different components, we perform an ablation study with the following variants of AAG,
	
	i)	Using only two basic views;
	
	ii)	Replacing the auxiliary view with a basic view;
	
	iii) Replacing CosineLR schedule with StepLR schedule;
	
	iv)	Replacing GNT-Xent loss with NT-Xent loss;
	
	v)	Using RandAugment (2 random operations with a magnitude of 10) as the auxiliary augmentation approach. 
	
	Table~\ref{tab_ab1} displays the $k$NN evaluation results on CIFAR10 and reveals that each component of AAG has a positive effect on improving the model accuracy. Among these variants, auxiliary data augmentation contributes most to the performance gain. It can be also observed that using policies of RandAugment achieves a close performance compared with that of using AutoAugment, indicating that the AAG method is relatively robust to the augmentation policies used to generate the auxiliary view. Thus, it is applicable to various datasets where the optimal policies are unavailable.  
	
	%. It means that our auxiliary data augmenation can be applied to more datasets directly without extra searching process.
	\begin{table}[H]
		\centering
		\caption{Ablation study on CIFAR10.}
		\label{tab_ab1}
		\begin{tabular}{lc}
			\toprule
			Network & Accuracy \\
			\midrule
			Full Model & \textbf{88.3} \\
			Two Basic Views & 85.8 \\
			Three Basic Views & 86.9 \\
			with StepLR & 87.1 \\
			with NT-Xent Loss & 86.8 \\
			with RandAugment & 88.1 \\
			\bottomrule
		\end{tabular}
	\end{table}
	
	\section{Conclusion}
	In this paper, we focus on the augmentation-based self-supervised learning and develop a new method called \textit{AAG}, which contains a new auxiliary augmentation scheme and a new GNT-Xent loss. The former introduces an auxiliary view in addition to the basic views to enhance the diversity of views and increase the number of data samples within a batch in the meantime. And the latter aims to achieve stable and efficient training. Both of these two components show advantages over their counterpart methods in the experiments. AAG improves the overall performance and works well in the condition of a small batch size which reduces space consumption, showing its great potential for unsupervised learning in computer vision tasks. %fWe hope our findings and explorations can be of benefit to help better understand the insights of self-supervised learning and inspire future research on large-scale datasets or related areas.

	\bibliographystyle{aaai21}
	\bibliography{references}
	
\end{document}